\documentclass[nohyperref]{article}

\usepackage{microtype}
\usepackage[pdftex]{graphicx}
\usepackage{subfigure}
\usepackage{booktabs} 

\usepackage[backref=page]{hyperref}

\usepackage[accepted]{icml2022}

\usepackage{amsmath}
\usepackage{amssymb}
\usepackage{mathtools}
\usepackage{amsthm}
\usepackage{xcolor}
\usepackage[capitalize,noabbrev]{cleveref}

\theoremstyle{plain}

\theoremstyle{definition}

\theoremstyle{remark}

\usepackage[textsize=tiny]{todonotes}

\newcommand{\dens}{\emph{density}}
\newcommand{\covr}{\emph{coverage}}

\newcommand{\Dens}{\emph{Density}}
\newcommand{\Covr}{\emph{Coverage}}
\icmltitlerunning{}

\begin{document}

\twocolumn[
\icmltitle{Detecting Adversarial Examples in Batches - a geometrical approach}



\icmlsetsymbol{equal}{*}

\begin{icmlauthorlist}
\icmlauthor{Danush Kumar Venkatesh}{yyy,xxx}
\icmlauthor{Peter Steinbach}{yyy}
\end{icmlauthorlist}

\icmlaffiliation{yyy}{Helmholtz-Zentrum Dresden-Rossendorf\\
Dresden, Germany}
\icmlaffiliation{xxx}{Technishe Universitat Bergakademie\\
Freiberg, Germany}

\icmlcorrespondingauthor{Danush Kumar Venkatesh}{ d.venkatesh@hzdr.de}
\icmlcorrespondingauthor{Peter Steinbach}{p.steinbach@hzdr.de}

\icmlkeywords{Machine Learning, ICML}

\vskip 0.3in
]



\printAffiliationsAndNotice{}  

\begin{abstract}
   Many deep learning methods have successfully solved complex tasks in computer vision and speech recognition applications. Nonetheless, the robustness of these models has been found to be vulnerable to perturbed inputs or adversarial examples, which are imperceptible to the human eye, but lead the model to erroneous output decisions. %
   In this study, we adapt and introduce two geometric metrics, \dens{} and \covr{}, and evaluate their use in detecting adversarial samples in batches of unseen data. We empirically study the metrics using MNIST and two real-world biomedical datasets from MedMNIST, subjected to two different adversarial attacks. Our experiments show promising results for both metrics to detect adversarial examples. We believe that his work can lay the ground for further study on these metrics' use in deployed machine learning systems to monitor for possible attacks by adversarial examples or related pathologies such as dataset shift.
\end{abstract}
\section{Introduction}
\label{sec:intro}


Neural networks (NN) are machine learning algorithms that have been demonstrated to achieve human-level performance in computer vision (e.g. deep neural network architectures, DNNs), in speech recognition \cite{hinton2012deep} and many other tasks of academic, industrial or cultural value. However, studies have shown that the stability (or robustness) of such models can be distorted by adding small human imperceptible perturbations to the input samples \cite{szegedy2013intriguing, goodfellow2014explaining} so that predictions by these systems are misguided. Such imperceptibly perturbed samples are termed as \emph{adversarial samples} (AEs) and can be crafted by \emph{adversarial attacks} \cite{brendel2017decision}. 
This unenviable property of the NNs causes major security concerns with regard to deployment in real-world applications like medical imaging \cite{lundervold2019overview} or self-driving cars \cite{evtimov2017robust} to name a few. 


\paragraph{Our Contributions} 

In this paper, we propose the use of \dens{} and \covr{} \cite{naeem2020reliable} as metrics for detecting AEs in batches of images. We have adopted these two metrics from the field of generative modeling and quality assessment of generative adversarial networks (GANs). 

In detail, we contribute the following aspects to the literature:

\begin{itemize}
\item We adopted \dens{} and \covr{} as metrics and propose a \emph{model-agnostic} method to detect adversarial samples in batches of images. 
\item We empirically demonstrate the capacity of the metric using standard and real-world datasets subjected to both white- and black-box attacks aligned to recommendations in \cite{carlini2017adversarial}.
\item We study the root cause of \dens{} and \covr{} curves obtained from the aforementioned experiments.
\item We have improved upon the computational efficiency of the proposed metrics.
\end{itemize}

Our work aspires to remain computational efficient and useful for practitioners. In other words, our approach only requires practical applications to "train" a k-nearest neighbor tree on the used training set of images or affiliated NN embeddings obtained from domain-specific application, e.g. NN derived image classification. Due to the availability of performant open-source libraries such as \texttt{faiss} \cite{JDH17} we consider this to be a minimal burden for practical concerns.

\section{Related Work}

In this section, we provide a short summary of AE detection techniques and provide some context to image quality metrics from a generative modelling perspective in order to provide context for \dens{} and \covr{}. 




\paragraph{Detecting Adversarial Examples} 

With a statistical approach, \cite{grosse2017statistical} proposed a model agnostic and kernel-based two-sample test. The mean discrepancy distance metric \cite{gretton2012kernel} was used to detect AEs from clean samples. Detectors based on kernel density and Bayesian uncertainty estimates were presented to distinguish AEs from noisy and clean samples \cite{feinman2017detecting}. A manifold learning strategy based on expansion models was used to study the local dimensionality of adversarial regions. Local Intrinsic Dimensionality (LID) estimates were used further to distinguish AEs \cite{ma2018characterizing}. Based on the assumption that the feature space of models can be fitted to class conditional Gaussian distributions by Gaussian discriminant analysis, a confidence score was suggested to detect out-of-distribution and AEs based on Mahalanobis distance \cite{lee2018simple}. A k-nearest neighbor(k-NN) model was fit to the feature space of DNNs utilizing influence function score \cite{cohen2020detecting}. Similarly, neighbor context encoder (NCE), a transformer-based detector, was proposed by \cite{mao2020learning}. \cite{li2017adversarial} built a cascade classifier to distinguish AEs using the PCA of samples from each output layer of DNNs. For a more detailed review of the field, we refer to \cite{aldahdooh2022adversarial}.


\paragraph{GAN image generation quality metrics} 

The qualitative performance of a GAN model is commonly assessed with respect to two main aspects: fidelity (quality) and diversity (variability) of the generated (fake) images with respect to the training data set. Multiple metrics have been presented to evaluate the models based on these aspects. More recently, two separate metrics, precision and recall, were formulated based on the uniform density assumptions of real and fake data distributions \cite{sajjadi2018assessing}. Further \cite{kynkaanniemi2019improved} proposed improved precision (IP) and recall (IR), wherein the probability density functions were used to cautiously construct the manifolds using the \emph{k}th nearest neighbor. Certain drawbacks of IP\&IR such as outlier susceptibility or the computational expense were overcome by \emph{density} and \emph{coverage}, that was implemented as \texttt{prdc} \cite{naeem2020reliable}.\\

\section{Adversarial example detection}

In this work, we focus on image classification and attacking such a trained model to produce AEs in an untargeted fashion, i.e., the model predicts any other class than the ground truth. In this section, we introduce \dens{} and \covr{} mathematically. 

Let $\mathcal{P}$(X) be the distribution of the dataset images and $\mathcal{Y}$ be the corresponding class labels. Following the underlying assumptions from \texttt{prdc}, we sample two different distributions, the model dataset $\mathcal{P}(X_{m})$ and a hold-out set, the validation dataset $\mathcal{P}(X_{v})$. Let the labeled set, $S_{m} = (x_{i}, y_{i})^{m}_{i=1}$ be sampled from $\mathcal{P}(X_{m})$ and  $S_{v} = (x_{j}, y_{j})^{n}_{j=1}$ from $\mathcal{P}(X_{v})$, with \emph{m} and \emph{n} the number of samples in model and validation dataset respectively. Let $f$ be a model trained on $\mathcal{P}(X_{m})$. The model $f$ is then attacked using $S_{v}$ to generate the adversarial sample $x_{a}$ with the distribution $\mathcal{P}(X_{a})$.

\subsection{Density} \label{den}

In the context of generative modeling, density quantifies the quality and coverage assesses the variability of the generated (fake) images. These metrics, have been developed with improved precision and recall(IP\&IR) as their backbone. Let $\mathcal{R}$ and $\mathcal{Q}$ be reference and query distributions, then IP\&IR is computed by constructing the manifolds $\Phi_{r}$ and $\Phi_{q}$, one for each distribution, with the hypersphere radii estimated by the \emph{k}-nearest neighbor. The binary decision of whether $q_{i}$ sampled from query distribution $\mathcal{Q}$ lies in the manifold $\Phi_{r}$ constitutes precision, whereas recall measures the portion of $\mathcal{R}$ falling inside $\Phi_{q}$. \Dens{} is computed by normalizing the constructed manifold and counting the number of reference manifold spheres containing $q_{i}$. It is defined as
\begin{equation}
    \label{eq:den}
	\text{density} = \frac{1}{kS} \sum_{j=1}^{S} \sum_{i=1}^{T} 1_{q_{j} \in \mathcal{M}(r_{1},\cdots,r_{S})}
\end{equation}
where \emph{S} and \emph{T} are the number of reference and query samples, $1_{(.)}$ the binary decision function and $\mathcal{M}$ the defined reference manifold.

\subsection{Coverage}

\Covr{} computation focuses on constructing manifolds only for the reference samples in contrast to IR and measures the fraction of these manifolds containing at least one query sample. \Covr{} is defined as
\begin{equation}
    \label{eq:cov}
   \text {coverage}=\frac{1}{T} \sum_{i=1}^{T} 1_{\exists j \text { s.t. } q_{j} \in \mathcal{M}(r_{1},\cdots,r_{S})}
\end{equation}
\Dens{} remains unbounded, that is, it can be larger than $1$, whereas \covr{} is bounded between $0$ and $1$. The manifolds are defined by
\begin{equation}
    \label{eq:manifold}
    \mathcal{M}\left(r_{1}, \cdots, r_{S}\right) = \bigcup_{i=1}^{S} \mathcal{H}\left(r_{i}, \operatorname{nn}_{k}\left(r_{i}\right)\right)
\end{equation}
where $\mathcal{H}(r_{i},\Tilde{r})$ is the sphere around $r_{i}$, with the radius $\Tilde{r}$ estimated by $\operatorname{nn}_{k}\left(r_{i}\right)$ that denotes the \emph{k} nearest neighbor to $r_{i}$, expressed in $\mathbf{R}^{d}$. The computation of these manifolds exposes a severe runtime overhead in the original implementation of \texttt{prdc}. We overcome this problem using the tree structure of \texttt{faiss} \cite{JDH17} to find the nearest neighbor to a given data point and efficiently construct the hyperspheres. \cref{runtime} is referred for related analysis. 

\section{Experiments}

We briefly explain the datasets and the adversarial attacks which were explored to quantify the proposed metric.

\subsection{Datasets}

The study was performed on the standard MNIST dataset \cite{lecun1998gradient} and also on 2D real-world biomedical datasets from MedMNIST \cite{medmnistv2}. The datasets are described as follows, \\ 

\textbf{MNIST}: The dataset consists of $70000$ grayscale images with $10$ classes corresponding to handwritten digits $0$ to $9$. The dataset was split into a model dataset size of $67900$ images and a validation dataset of $2100$ images. \\ 

\textbf{PathMNIST}: This dataset consists of $107,180$ pathological tissue images consisting of $9$ classes describing different pathological tissue types. A total of $5359$ images were used to form the validation dataset, and the remaining was the model dataset. This dataset is encoded as a set of RGB images \cite{pathmnist}. \\ 

\textbf{OrganMNIST}: $58,850$ abdominal CT scan grayscale images corresponding to $11$ different classes are contained in this dataset \cite{organmnist}. A validation set size of $4708$ images was chosen for our study.\\ 

\begin{figure}
    \centering
    \begin{tabular}{c}
    \subfigure[MNIST]{\includegraphics[width=\linewidth]{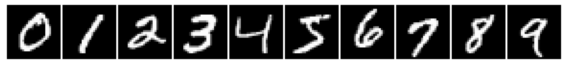}} \\
    \subfigure[PathMNIST]{\includegraphics[width=\linewidth] {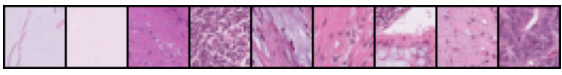}} \\
    \subfigure[OrganMNIST]{\includegraphics[width=\linewidth] {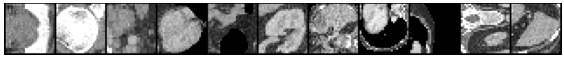}}
    \end{tabular}
    \caption{The images of the different dataset. Each image represents a unique class in the dataset.}
    \label{fig:dataset_imgs}
\end{figure}

All the images were of $28$x$28$ pixels. Each dataset was split using a stratification strategy. The proportional values are available in \cref{model_info}. For illustration, we show a subset of each dataset in ~\cref{fig:dataset_imgs}. 


\subsection{Procedure for metric computation}

A necessary preliminary step for the computation of the metric and the interpretation of our results is the dataset curation. We conducted this as follows:

\begin{enumerate}
\item \textbf{Dataset split}: The dataset of interest is split into $2$ different sets, a set to train the model, and a hold-out dataset. We mention these splits as the model dataset and the validation dataset. The model dataset is used to train the CNN model, and the validation dataset is used exclusively to perform the attack and generate adversarial samples.
\item \textbf{Generate adversarial samples}: Attack the trained model using the validation dataset and generate the \emph{validation adversarial samples}
\end{enumerate}
However, if adversarial samples are already present, step 2 can be bypassed. 

The procedure to differentiate and detect the adversarial samples using the proposed metrics is as follows:
\begin{enumerate}
\item \textbf{Reference \dens{}/\covr{}}: compute the metrics between the  \emph{model dataset} and the \emph{validation dataset}\- which serves as reference value before the model is attacked.
\item \textbf{Adversarial \dens{}/\covr{}}: compute the metrics between the \emph{model dataset} and the \emph{validation adversarial samples}.
\end{enumerate}
The reference manifold is always constructed on the model dataset($\Phi_{m}$) and the query samples constitute either the validation dataset or the generated AEs to compute the reference and adversarial metric respectively. 

In our setup, \dens{} and \covr{} are metrics to describe the geometric distribution of samples from the holdout set (e.g. a query batch) with respect to the distribution of the data used for training. Along this line of thought, our method can be applied on any input data that represents a query batch uniquely in an appropriate n-dimensional space. In other words, these metrics can be computed on the entire benign or malignant query image batch. Alternatively, both metrics can also be calculated from the embedded space of logits of a trained classifier or autoencoder as obtained from the query image. In the following, we will report the results obtained for the entire images only. Interested readers curious about the outcomes in the embedded space are kindly deferred to \cref{Met_analysis}.

\subsection{Adversarial attacks}

We focus our study on two attacks namely, Fast gradient sign method (FGSM)~\cite{goodfellow2014explaining} and the Boundary attack~\cite{brendel2017decision}. We use these two extremes to probe the rich space of possible attacks already published. As our method is independent of the attack, we consider it highly likely that our reported results map to other attacks too. 

\paragraph{FGSM} A type of white-box attack in which AEs are crafted by computing the gradient of the loss function for the given input image such that the loss increases along the steepest direction. In our study, it was meaningful to vary the magnitude of perturbation, $\epsilon$, in the range $[0, 1]$ in steps of $0.05$ as the accuracy of the models decreased to below-par values.

\paragraph{Boundary attack} In this algorithm, AEs are generated by reducing the distance between the original (non-adversarial) sample and adversarial target (i.e., a sample from a target class to be injected) by tracing along the decision boundary and making a step-update in combination with a suitable perturbation from a given distribution.

Foolbox \cite{rauber2017foolbox} was used to attack our trained models and to generate the AEs.

\subsection{CNN Model}

We designed two different model architectures for the classifier model on which the attack was performed: A $5$ layer model for the MNIST dataset. For both the MedMNIST datasets, an architecture of $5$ convolutional layers followed by $3$ fully-connected layers was created. Both MedMNIST models are only different in the number of channels for the input layer. The model details are provided in \cref{model_info}.

To define the uncertainties of the metrics for the attacks, uncertainty estimates were obtained for both reference and adversarially altered metrics. For this, the respective image datasets were split into smaller batches of $100$ images each. Then, any metric of interest was collected for each of these sub-batches. From the ensemble of measurements obtained this way, we compute the $0.25$, $0.5$, and $0.75$ quantiles of any metrics of interest for this study. 

For the FGSM attack, we use $20$ values for epsilon, and both metrics are computed for each of these values. In the case of the Boundary attack, we also examine the metrics by varying the batch size of the images depending on the datasets. Throughout this study, $k$ as of~\cref{eq:den} and ~\cref{eq:cov} was maintained at a value of $5$ in any nearest neighbor query performed.
The code and the implementation of experiments are provided in~\cite{n2gem_repo}.

\section{Results \& Discussion}

\begin{figure}[t]
    \setlength\tabcolsep{2pt}
    \centering
    \begin{tabular}{p{0.25\linewidth} p{0.10\linewidth} p{0.1\linewidth} p{0.1\linewidth} p{0.15\linewidth} p{0.1\linewidth}}
    \toprule
         \textbf{Dataset} &
        \textbf{Original} &
         \multicolumn{3}{c}{\textbf{FGSM}} &
         \textbf{Bdy.}\\
         \midrule
         \textbf{MNIST} & \includegraphics[width=1\linewidth, height=0.05\textwidth] {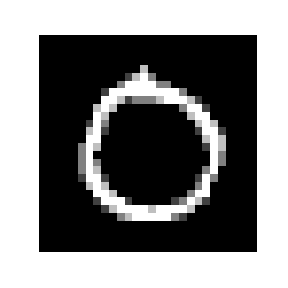} & \includegraphics[width=1.\linewidth, height=0.05\textwidth]{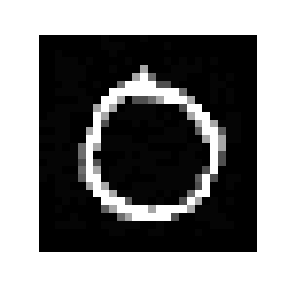} & \includegraphics[width=1.\linewidth, height=0.05\textwidth]{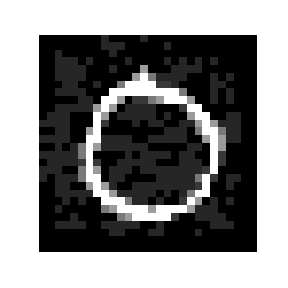} & \includegraphics[width=0.75\linewidth, height=0.05\textwidth]{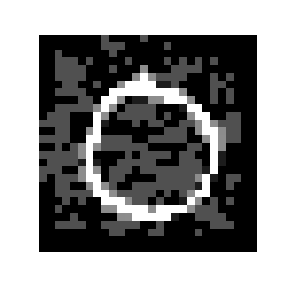} & \includegraphics[width=1\linewidth, height=0.05\textwidth]{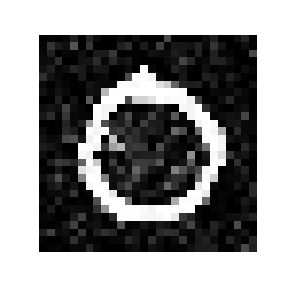} \\[-0.5em]
         & \includegraphics[width=1\linewidth, height=0.05\textwidth]{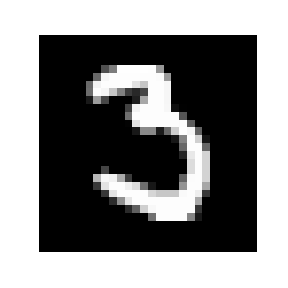} & \includegraphics[width=1\linewidth, height=0.05\textwidth]{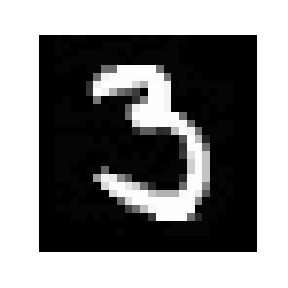} & \includegraphics[width=1\linewidth, height=0.05\textwidth]{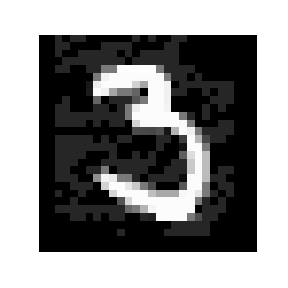} & \includegraphics[width=0.75\linewidth, height=0.05\textwidth]{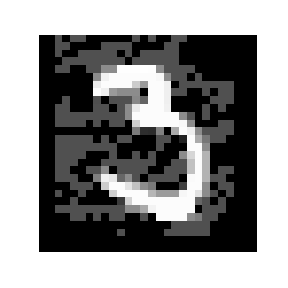} & \includegraphics[width=1\linewidth, height=0.05\textwidth]{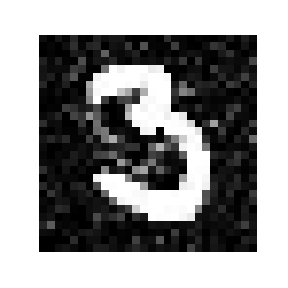}\\ \midrule
         \textbf{PathMNIST} & \includegraphics[width=1\linewidth, height=0.05\textwidth]{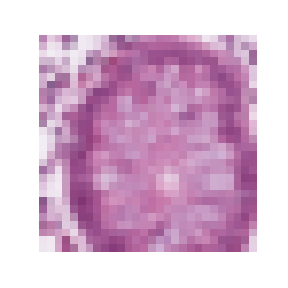} & \includegraphics[width=1\linewidth, height=0.05\textwidth]{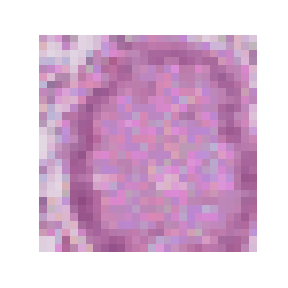} & \includegraphics[width=1\linewidth, height=0.05\textwidth]{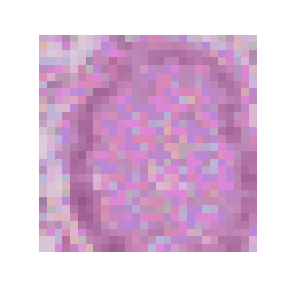} & \includegraphics[width=0.75\linewidth, height=0.05\textwidth]{images/adv_images/fgsm_path_2011_epi2.png} & \includegraphics[width=1\linewidth, height=0.05\textwidth]{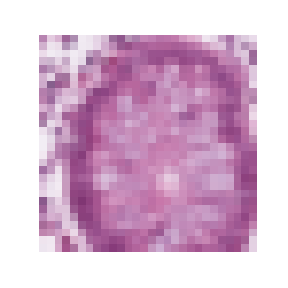}\\[-0.5em]
         & \includegraphics[width=1\linewidth, height=0.05\textwidth]{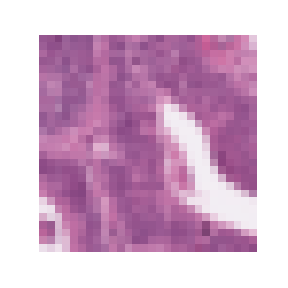} & \includegraphics[width=1\linewidth, height=0.05\textwidth]{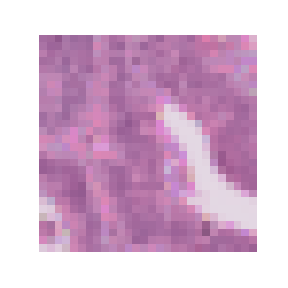} & \includegraphics[width=1\linewidth, height=0.05\textwidth]{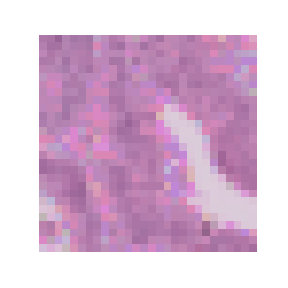} & \includegraphics[width=0.75\linewidth, height=0.05\textwidth]{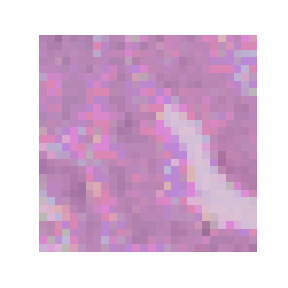} & \includegraphics[width=1\linewidth, height=0.05\textwidth]{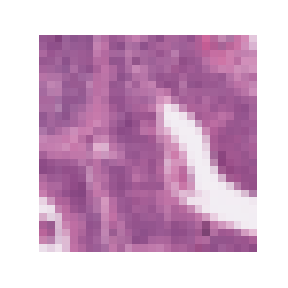}\\ \midrule
         \textbf{OrganMNIST} & \includegraphics[width=1\linewidth, height=0.05\textwidth]{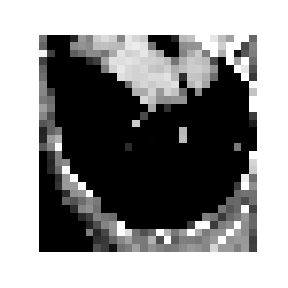} & \includegraphics[width=1\linewidth, height=0.05\textwidth]{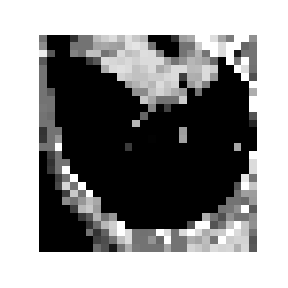} & \includegraphics[width=1\linewidth, height=0.05\textwidth]{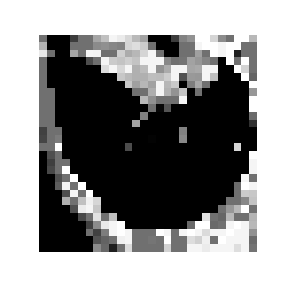} & \includegraphics[width=0.75\linewidth, height=0.05\textwidth]{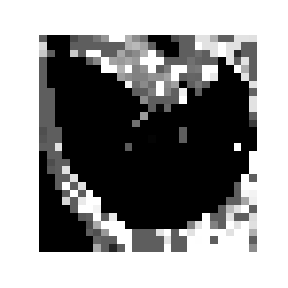} & \includegraphics[width=1\linewidth, height=0.05\textwidth]{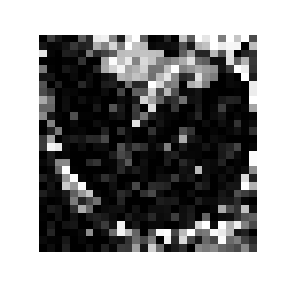}\\[-0.5em]
         & \includegraphics[width=1\linewidth, height=0.05\textwidth]{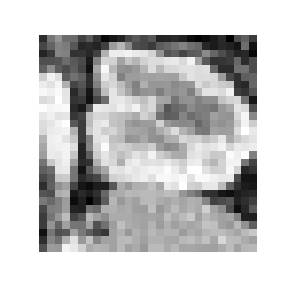} & \includegraphics[width=1\linewidth, height=0.05\textwidth]{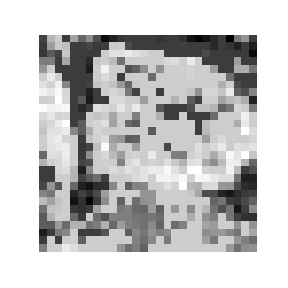} & \includegraphics[width=1\linewidth, height=0.05\textwidth]{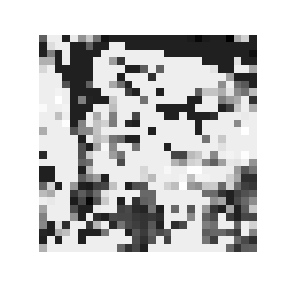} & \includegraphics[width=0.75\linewidth, height=0.05\textwidth]{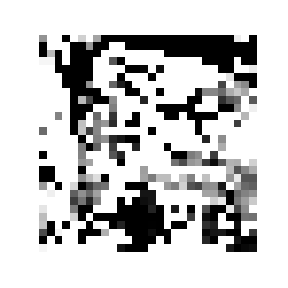} & \includegraphics[width=1\linewidth, height=0.05\textwidth]{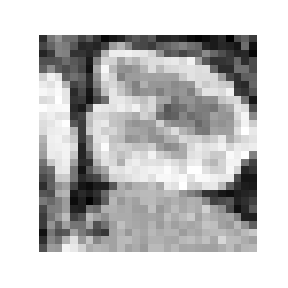}\\ 
         \small
         $\epsilon_{\text{FGSM}}$ & & \centering $0.3$ & \centering $0.75$ & \centering $1.0$ & \\
         
         \bottomrule
         
    \end{tabular}
    \caption{Example images from the original datasets as well as generated by two adversarial attacks from our classification models. The $2^{nd}$ column indicates the original (benign) samples. $3^{rd}$, $4^{th}$ and $5^{th}$ columns correspond to FGSM attack with $\epsilon_{\text{FGSM}}$ being set to $0.3,0.75$ and $1$ as indicated. The last column of images are AEs generated from the boundary attack.}
    \label{fig:adv_samples}
\end{figure}

We summarize and discuss the results of the experiments in this section. The CNN model trained for the MNIST dataset achieved an accuracy of $99.2~\%$. Similarly, the models for the medical images were trained to achieve an accuracy of $ 96~\%$  and $99.1~\%$ for PathMNIST and OrganMNIST, respectively. The model accuracy for biomedical images was on par with the benchmark results from \cite{medmnistv2}. The AEs generated for different datasets are shown in ~\cref{fig:adv_samples} to provide an overview and intuition.

 \subsection{FGSM attack} \label{fgsm}
\begin{figure*}
    \centering
    \subfigure[FGSM attack ]{\label{fig:fgsm_variation}{\includegraphics[width=0.45\linewidth] {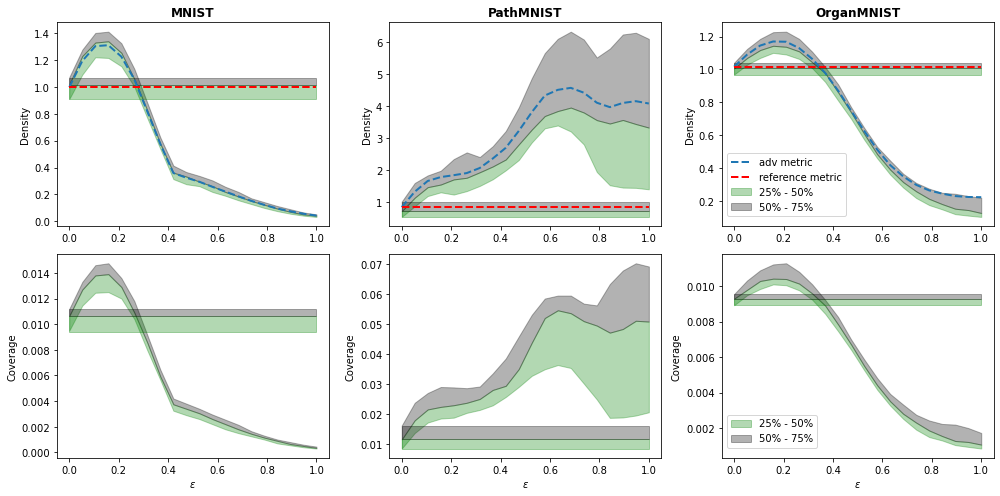}}}%
    %
  \hfill
    \subfigure[Boundary attack ]{\label{fig:bdy_variation}{\includegraphics[width=0.45\linewidth] {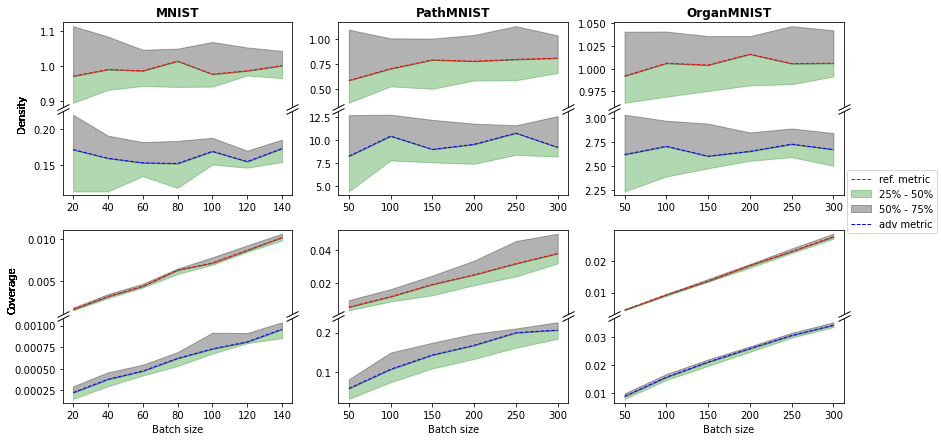}}}%
    
  \caption{The variation of \dens{} and \covr{} for the reference validation dataset as well as for AEs obtained from it. The red line corresponds to metrics obtained from the benign (reference) validation set. The blue line refers metrics obtained from malignant AEs derived from the same validation set. The green and gray bands indicate uncertainty intervals between $25\%-50\%$ and $50\%-75\%$ quantiles estimated from ensemble measurements respectively. Each column corresponds to the specified dataset as indicated. Rows report results for \dens{} or \covr{} respectively. \Dens{} and \covr{} from AEs are distinctly different from reference values.}
  \label{fig:met_var}
\end{figure*}

 The variation of \dens{} and \covr{} for the FGSM attack on the datasets is shown in ~\cref{fig:fgsm_variation}.
 
\paragraph{MNIST}%
The reference \dens{} for the MNIST dataset achieves a value of $0.997$ (horizontal line in~\cref{fig:fgsm_variation} $1^{st}$ row). After the attack, the adversarial \dens{} varies from the reference \dens{} for the complete range of epsilon values. It increases gradually with an increase in epsilon and crosses over towards small values of \dens{} at $\epsilon=0.38$, after which it gradually decreases further with the increase in epsilon. 
%
The reference \covr{} for the dataset has a value of $0.39$. Similar to adversarial \dens{}, the adversarial \covr{} for the MNIST dataset increases initially with a peak value of $0.46$, then crosses over the reference value at $\epsilon=0.28$ and decreases further towards smaller values.
\paragraph{MedMNIST}%
The \dens{} variation for the OrganMNIST dataset shows similar behavior to the MNIST dataset. The reference \dens{} has a value of $1.01$, and the adversarial \dens{} peaks at $1.17$ and reduces to $0.22$ for the maximum epsilon value. However, for the case of PathMNIST, a non-identical trend in the adversarial \dens{} was noticed. This dataset has a reference \dens{} value of $0.83$, and the adversarial \dens{} raises thereafter with the increase in epsilon, having the maximum value of $4.56$.
The behavior and variation in \covr{} for all three datasets appears similar to each other. In the case of OrganMNIST, the reference \covr{} had a value of $0.34$, and the adversarial \covr{} varies between $0.36$ and $0.003$. The adversarial \covr{} has a sharp peak at $\epsilon=0.18$ for PathMNIST dataset, whereas the reference \covr{} remained at $0.49$.%
\paragraph{Metric performance}
As described for each dataset, the reference \dens{} lies close to the value of $1$. This is an expected outcome, as it confirms the i.i.d assumption. It also indicates that the model dataset manifold $\Phi_{m}$  is closely packed without outliers along with the validation samples lying in the vicinity~\cite{naeem2020reliable}. From the ~\cref{fig:fgsm_variation}, it is evident that for the full range of epsilon values, both adversarial \dens{} and \covr{} exhibit a noticeable difference from the reference metric. For the MNIST and OrganMNIST datasets, the adversarial \dens{} increases in the region between $\epsilon=0$ and $\epsilon=0.38$, which is purely due to geometric reasons. Concerning the definition of the FGSM attack, the gradient tensor with respect to the input image is nudged in a direction with the scaling value of epsilon to create an adversarial sample. As the computation of \dens{} and \covr{} is based on the binary function, as seen in ~\cref{eq:den} and ~\cref{eq:manifold}, with the initial increase of epsilon, the generated AEs move into the multitude of model dataset manifolds $\Phi_{m}$, and this augments the number of hyperspheres $\mathcal{H}$ containing AEs thereby contributing to the initial increase of the metric. 

With higher epsilon values, the decreasing trend of the \dens{} curves underlines that the manifold, $\Phi_{m}$ and AEs become separated. The continuous increase of adversarial \dens{} in the case of PathMNIST revealed that the generated AEs are classified into a particular class with the increase in epsilon, thereby confining the AEs into a small region of hyperspheres and contributing to the \dens{} increase. As the value obtained for \covr{} depends on the number of samples considered for its computation, a variation in reference values \covr{} between the datasets is expected. From the \covr{} plots, the decreasing \covr{} values with the increase in epsilon indicate the reduction in the diversity of samples generated. That is, the generated AEs always tend to be classified into a particular class with the increase of epsilon. A visualization in a space of reduced dimensionality using \texttt{pymde} \cite{agrawal2021minimum} and an analysis of it are provided in ~\cref{Met_analysis}. 

\subsection{Boundary attack} \label{bb_atk}
The behaviour of \dens{} and \covr{} between the benign and AEs generated using the boundary attack for the datasets are shown in ~\cref{fig:bdy_variation}. 
\paragraph{MNIST}
For this dataset, the reference \dens{} lies between $0.9$ and $1.1$ for various batch sizes. The adversarial \dens{} values were comparatively low, within the range of $0.11$ and $0.22$. The reference \covr{} achieves a value of $0.14$ compared to the adversarial \covr{} value of $0.004$ for the entire batch of our validation dataset. An interesting result, as noticed in the $2^{nd}$ column of ~\cref{fig:bdy_variation}, is that \covr{} increases gradually with an increase in image batch sizes and finally reaches that of reference or adversarial values. %
\paragraph{MedMNIST}%
In the case of PathMNIST, an averse behavior of \dens{} to that of MNIST was noticed. The reference \dens{} has a value of $0.831$ and the adversarial \dens{} with the value of $10.12$ for the whole batch of validation dataset. A slight difference in values was measured in the OrganMNIST dataset,:with reference \dens{} lying between $0.96$ and $1.05$ compared to adversarial \dens{} values of $2.2$ and $3.1$. The \covr{} values for these datasets showed the same inclination as seen in the MNIST dataset: the gradual increase of \covr{} with the increase in batch sizes. For the entire batch of images, we obtained the reference \covr{} for PathMNIST at a value of $0.49$ and the adversarial \covr{} value of $0.65$. For the OrganMNIST, the reference \covr{} was at $0.34$ in meager contrast to adversarial \covr{} of $0.24$. %
\paragraph{Metric performance}%

Similar to the FGSM attack, the reference values of density and coverage remained the same. However, for the case of MNIST and PathMNIST datasets, a discernible difference was noticed between the reference and adversarial metrics. The progressive increase of \covr{} with an increase in the batch size reassures us of the dependency on the number of samples involved in its computation. The results also exhibit the distinctive capability of \dens{} and \covr{} to detect the adversarial samples. The variation in these metrics reveals that the pathological characteristics of AEs were more adverse than those crafted by the FGSM attack. %

\subsection{Admixture of adversarial samples}
In a real-world scenario, practitioners aspire to detect AEs in batches of samples to classify on a case by case basis at best. In order to learn, how precise \dens{} and \covr{} can be used for this purpose we intermixed AEs into subsets of benign (untouched) samples in proportionate quantities. We then computed our metrics of interest. The metrics variation for each of the dataset for the FGSM attack is shown in ~\cref{fig:advMix_variation}. The results of the similar analysis for the boundary attack is available in the supplementary material. 

As seen from these results, \dens{} and \covr{} show a distinctively progressive deviation from the full batch of benign samples (i.e. a deviation away from the reference metric), to that of the batch of AEs for the attack. These experiments further advocate that \dens{} and \covr{} can serve as a measure to detect AEs.


\begin{figure}[t]
    \centering
    \includegraphics[width=\linewidth] {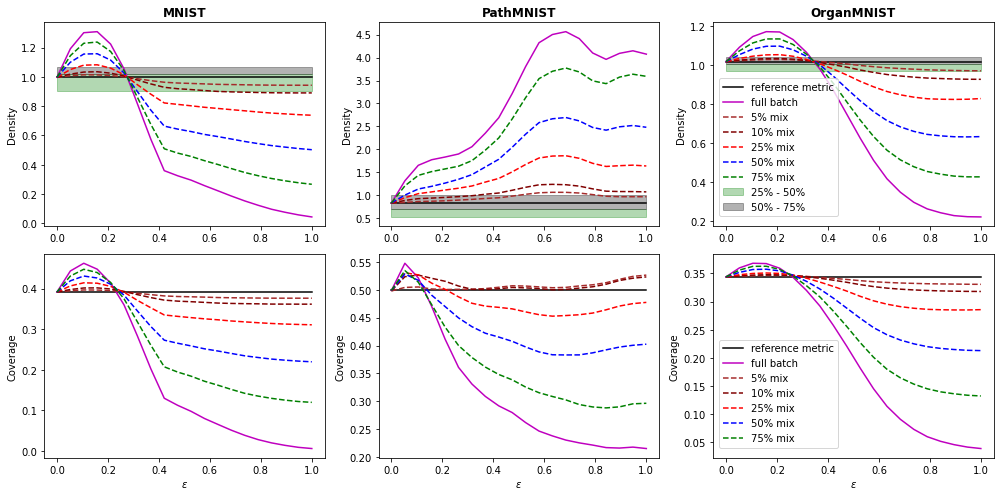}
    \caption{The proportionate mixture of adversarial samples and benign samples for the FGSM attack. \Dens{} and \covr{} for the entire batch of benign(reference) and adversarial samples are indicated by black and purple lines, respectively. The red, blue, and green lines indicate the variation of the metrics for $25\%$, $50\%$, and $75\%$ proportional mixture of AEs with benign samples. \Dens{} variation is specified in the first row, followed by \covr{} in the second row. Above values of 25\% of malignant sample admixture, \dens{} and \covr{} values start to diverge above dataset-specific thresholds.}
    \label{fig:advMix_variation}
\end{figure}


\subsection{Discussion} 
\label{dis}
An essential property of our proposed metrics is that they are model-agnostic. In that spirit, they can directly operate on the feature space or the data itself. This overlooks the development of auxillary models for AE detection. In contrast to the usefulness of \dens{} and \covr{} for detection of AEs, we see a few shortcomings from our analysis above: for one, the sensitivity of \covr{} to the number of samples; further the number of nearest neighbors used for computation, which is inherent to the mathematical formulation of \covr{} computation is another. 

\section{Conclusion}

We have adopted two metrics for quality of sample generation from generative modeling, i.e. \dens{} and \covr{}. We improved their design to make them of practical use in realistic runtime scenarios. We then studied how \dens{} and \covr{} behave in the face of AEs for classifying images from MNIST or from benchmark datasets in the medical domain, PathMNIST and OrganMNIST. Our experiments confirm that both \dens{} and \covr{} are susceptible to effects maliciously introduced in AEs which are imperceptible to humans. Moreover, we demonstrated that admixtures of AEs into batches of untouched benign images by $25\%$ or more distort the values of \dens{} or \covr{} in such a fashion, that it makes both metrics viable candidates to flag such batches of images as malignant in practice. We believe that this can and should be subject of further study. With a potential addition of statistical hypothesis tests to differentiate provided (unseen) samples from a known training set using \dens{} and \covr{}, our approach can be of substantial support in fraud detection and monitoring of adversarial attacks in deployed machine learning systems. Apart from this, applications in detecting dataset shift or similar pathologies specific to a presented dataset in comparison to the training set should also be considered in future work.




{\small
\bibliographystyle{icml2022}
\bibliography{bibfile}
}

\newpage
\appendix
\onecolumn
\section{Adversarial attacks}
We followed the categorization of adversarial attacks alongside \cite{biggio2018} into:
\begin{itemize}

\item \textbf{White-box-attacks}: The attacker has complete information regarding the target model, including the model parameters and the architecture. AEs are crafted by formulating an optimization problem.
\item \textbf{Gray-box-attacks}: In this category, the attacker has information regarding the model architecture or the training data used to devise the parameters but no direct access to the learned parameters.
\item \textbf{Black-box-attacks}: Adversarial examples are crafted by a query to the trained model and observing the predicted label. 
\end{itemize}

\paragraph{FGSM} The adversarial sample for the attack is formally defined as,
\begin{equation}
    x_{a} = x + \epsilon * sign(\nabla_{x}J(\theta,x,y))
\end{equation}
where $\epsilon$ is the magnitude of perturbation, $\theta$ are the network parameters, and $J(\theta,x,y)$ is the cost or loss function of the trained model.\\ \\

\paragraph{Boundary attack} This is a rejection sampling algorithm to construct AEs. An adversarial target (i.e. a sample from a target class to be injected) is used as the starting point in this attack. A random perturbation drawn from the Gaussian distribution is added to the benign source image iteratively. Then, a step-update along the decision boundary between the original (non-adversarial) sample and the AE is performed such that the distance between them is minimized. Since this attack has information only about the predicted class label, multiple queries to the model have to be made before convergence, i.e. before a misclassification based on the AE is performed by the victim model. The AEs are generated by the minimization $\lVert x - x_{a} \rVert_{2}^{2}$, such that the new adversarial is defined by $x_{a}=x^{s-1}_{a}+\eta_{s}$, where $\eta_{s}$ is the random perturbation from a suitable distribution for a particular step $s$.
\section{Model information} \label{model_info}
The model architectures for the MNIST and the MedMNIST dataset are shown in ~\cref{tab:mnist_cnn} and ~\cref{tab:medmnist_cnn} respectively. Throughout the model, the kernel size was maintained at with radius $3$ along with a drop-out rate of $0.5$. To have a stabilizing effect on learning and to speed up the learning process, \emph{batch normalization} was introduced in the convolution layers. Except for the last layer, \emph{ReLU} was applied as the activation function and the \emph{softmax activation} for the last layer. The Adam optimizer and negative log-likelihood loss function was used. To estimate optimizer hyper parameters a $1$ cycle policy was employed, wherein the learning rate is varied for each mini-batch to train the model. The optimal learning rate for the dataset was found and the model was fine-tuned to achieve the best possible outcome. All models were trained using FastAI \cite{howard2020deep}. The split of samples into the model dataset and validation set for the datasets are indicated in ~\cref{tab:dataset_split}. 
\begin{table}[ht]
\centering
\setlength{\tabcolsep}{5pt}
\begin{tabular}{c p{0.5\textwidth}  c}
    \toprule
    Layer No. & Layer information & Description \\ \midrule
        1 & Conv2D + ReLU + MaxPool & 16 filters \\ \midrule
        2 & Conv2D + ReLU + MaxPool + Dropout & 32 filters \\ \midrule
        3 & Flatten & 800 neurons \\ \midrule
        3 & Dense + ReLU + Dropout & 512 neurons \\ \midrule
        4 & Dense + ReLU & 64 neurons \\ \midrule
        5 & Dense + Softmax & 10 classes \\ \bottomrule
\end{tabular}
\caption{The MNIST classifier architecture}
\label{tab:mnist_cnn}
\end{table}

\begin{table}[ht]
\centering
\setlength{\tabcolsep}{5pt}
\begin{tabular}{c p{0.5\textwidth} c} 
    \toprule 
    Layer No. & Layer information & Description \\ \midrule
        1 & Conv2D + Batchnorm + ReLU & 32 filters \\ \midrule
        2 & Conv2D + Batchnorm + ReLU + MaxPool & 32 filters \\ \midrule
        3 & Conv2D + Batchnorm + ReLU & 64 filters \\ \midrule
        4 & Conv2D + Batchnorm + ReLU & 64 filters \\ \midrule
        4 & Conv2D(padding=$1$) + Batchnorm + ReLU + MaxPool & 64 filters \\ \midrule
        5 & Dense + ReLU  & 128 neurons \\ \midrule
        6 & Dense + ReLU & 64 neurons \\ \midrule
        7 & Dense + Softmax & *classes \\ \bottomrule
\end{tabular}
\caption{The MedMNIST classifier architecture. *classes$=9$ and $11$ for PathMNIST and OrganMNIST respectively}
\label{tab:medmnist_cnn}
\end{table}

\begin{table}[ht]

\centering
\setlength{\tabcolsep}{5pt}
\begin{tabular}{c p{0.22\textwidth }p{0.22\textwidth} p{0.22\textwidth}} 
    \toprule 
    Dataset & Total no$.$ of samples & No$.$ of samples in model dataset & No$.$ of samples in validation set \\ \midrule 
    MNIST & $70000$ & $67000$ & $2100$ \\ \midrule
    PathMNIST & $107180$ & $101821$ & $5359$ \\ \midrule
    OrgMNIST & $58850$ & $54142$ & $4708$ \\ \bottomrule
\end{tabular}
\caption{The number of samples considered in each of the datasets}
\label{tab:dataset_split}
\end{table}

\clearpage
\section{Metrics analysis} \label{Met_analysis}
\begin{figure}[ht]
    \centering
    \includegraphics[width=\textwidth]{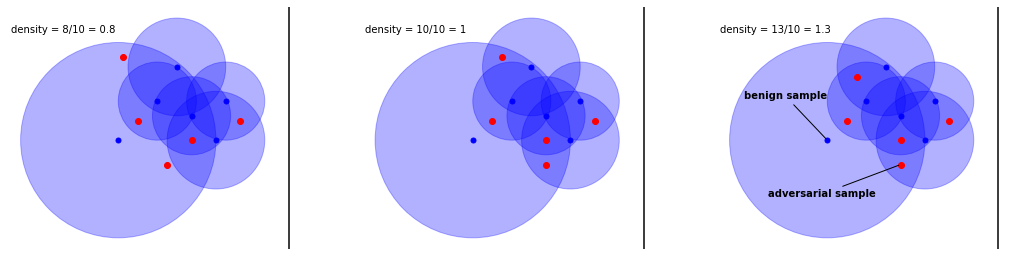}
    \caption{The graphical interpretation of the metric. The blue circles indicate the manifold of the benign samples and the red points correspond to AEs. The manifolds are formed with $k=2$.}
    \label{fig:den_exp}
\end{figure}

 The value of density and coverage depends on the output of the binary function—a short explanation of how density increases or decreases are shown in \cref{fig:den_exp}. 

The plot from left to right can be associated with the metrics variation for initial values of epsilon. As epsilon increases for the FGSM attack, the AEs (\emph{red points}) move into the benign sample hyperspheres (\emph{blue spheres}) and augment the number of hyperspheres for the AE. This contributes to the increase of the metric. However, with higher epsilon values, the AEs have moved away from the benign sample manifolds, thereby decreasing the metrics. For the PathMNIST dataset, an adverse behavior of adversarial density is noticed. As epsilon increases, the AEs initially move into the benign hyperspheres and get accommodated thereafter; the AEs for this dataset lies close to the benign samples. This is reflected in the continuous increase of density. \\ 

\begin{figure}[ht]
    \centering
    \includegraphics[width=0.8\textwidth, height=0.6\textwidth]{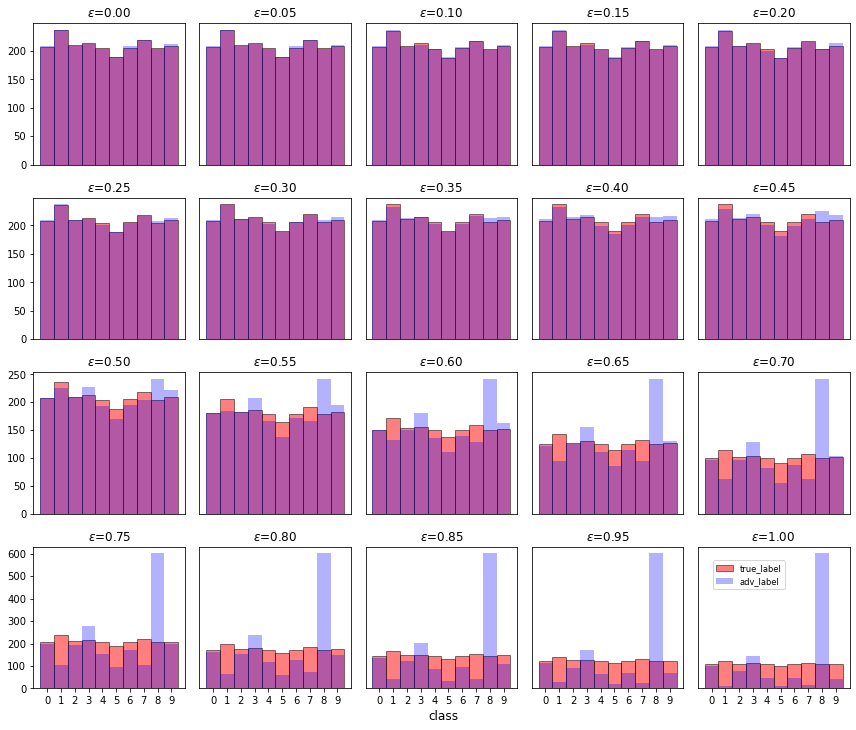}
    \caption{The variation between the true and predicted adversarial labels of the MNIST dataset subjected to FGSM attack for different value of epsilons.}
    \label{fig:fgsm_mnist_hist}
\end{figure}

\begin{figure}[ht]
    \centering
    \includegraphics[width=0.8\textwidth, height=0.6\textwidth]{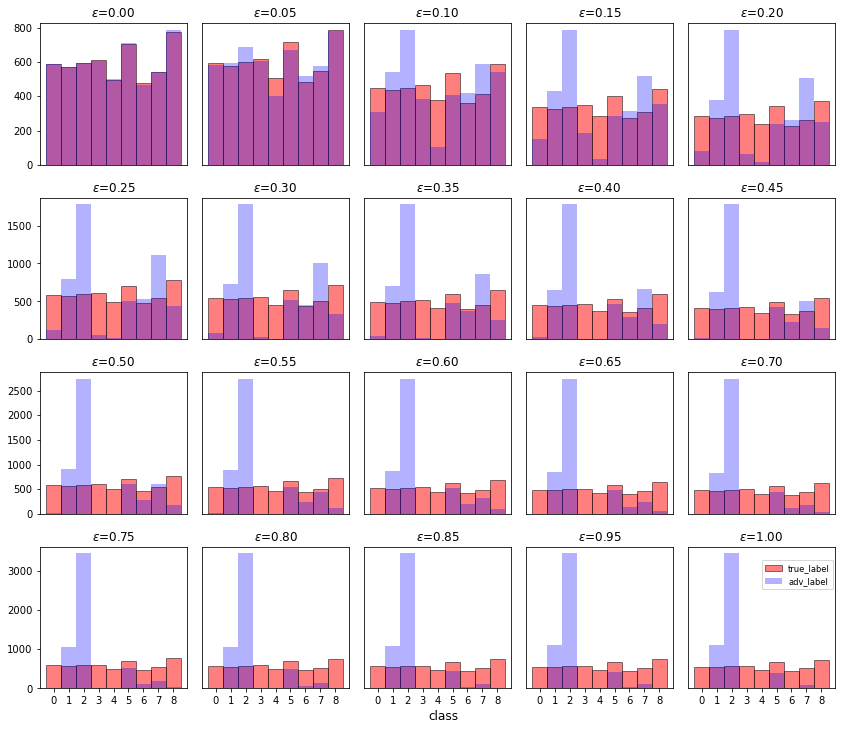}
    \caption{The variation for the PathMNIST dataset between the true and predicted adversarial labels for different values of epsilons. This illustrates how the attack changes the predicted class labels.}
    \label{fig:fgsm_pathmnist_hist}
\end{figure}

\begin{figure}[ht]
    \centering
    \includegraphics[width=0.8\textwidth, height=0.6\textwidth]{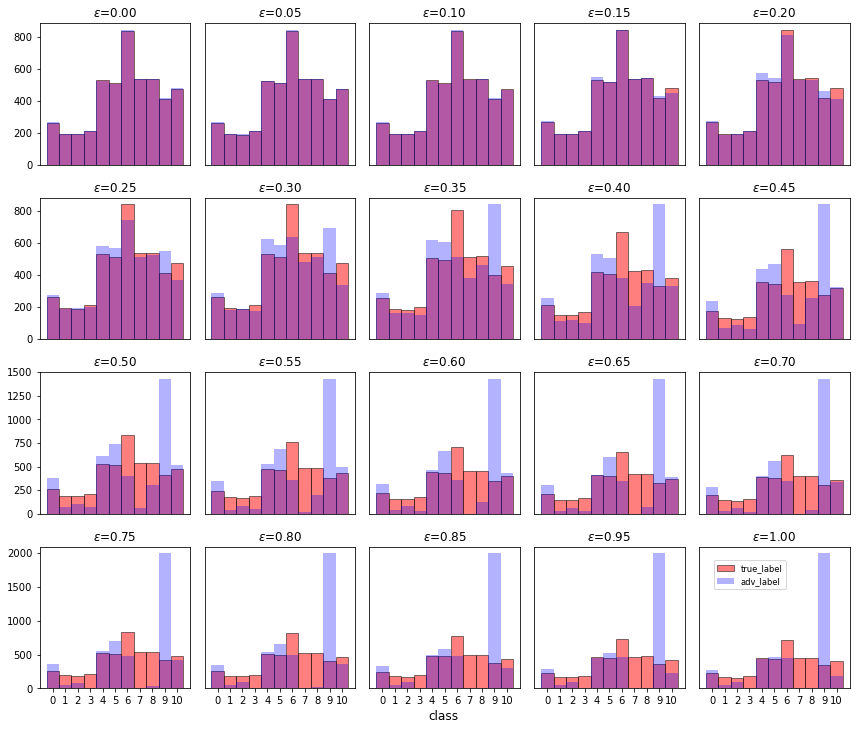}
    \caption{Variation between the true and predicted adversarial labels of the OrganMNIST dataset subjected to FGSM attack for different values of epsilons. This illustrates how the attack changes the predicted class labels.}
    \label{fig:fgsm_organmnist_hist}
\end{figure}

\begin{figure}
     \centering
        \subfigure[MNIST]{\label{fig:mnist_pymde}{\includegraphics[width=\textwidth]{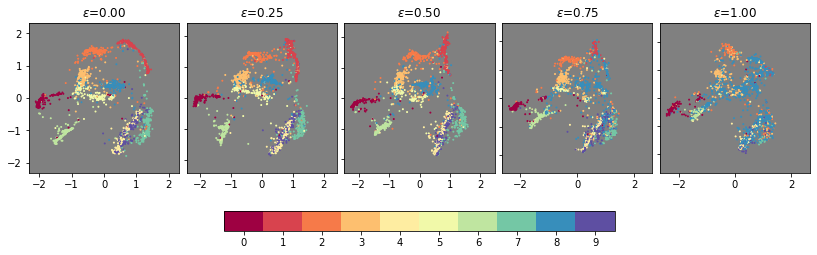}}}%
        \vfill
        \subfigure[PathMNIST]{\label{fig:pathmnist_pymde}{\includegraphics[width=\textwidth]{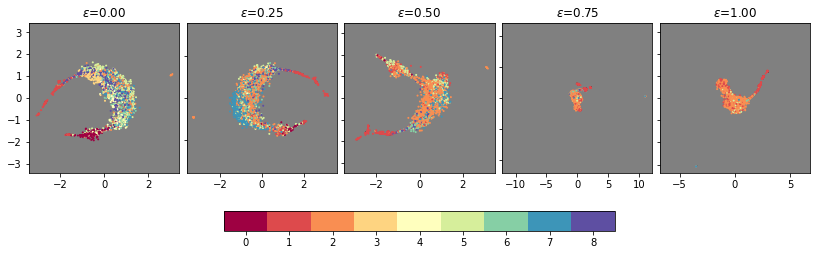}}}%
        \vfill
        \subfigure[OrganMNIST]{\label{fig:orgmnist_pymde}{\includegraphics[width=\textwidth]{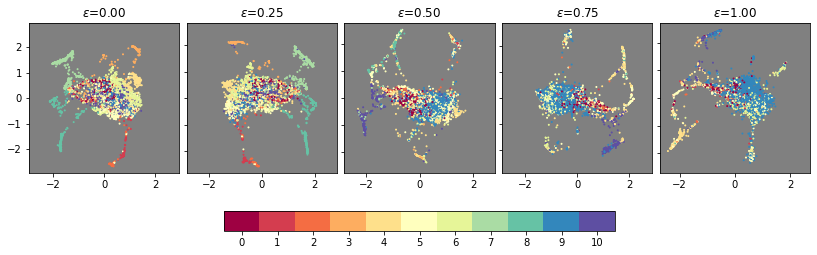}}}%
        \caption{Reduced dimensionality representation of the adversarial samples for each of the dataset. The colorbar indicates the label of each class in the dataset.}
    \label{fig:pymde_data}
\end{figure}

\begin{figure}
     \centering
     
        \subfigure[MNIST]{\label{fig:rdr_mnist}{\includegraphics[width=\textwidth] {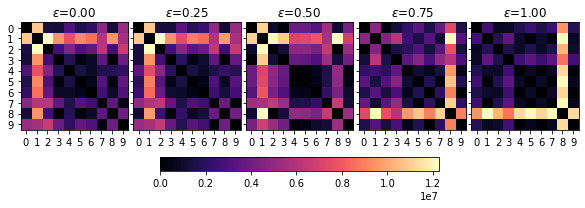}}}%
        \vfill
        \subfigure[PathMNIST]{\label{fig:rdr_pathmnist}{\includegraphics[width=\textwidth] {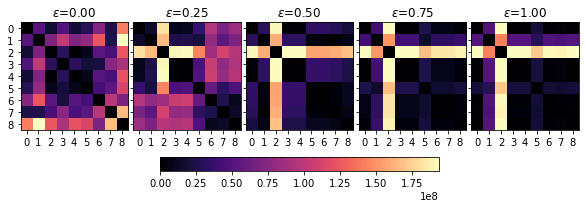}} }
        \vfill
        \subfigure[OrganMNIST]{\label{fig:rdr_orgmnist}{\includegraphics[width=\textwidth]{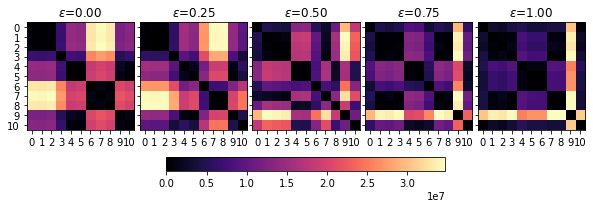}}}
    \caption{The difference in volume between the class clusters for each epsilon. The plots are symmetric, with yellow indicating a distinct difference between cluster volume and black no difference at all.}
   \label{fig:rdr_data}
\end{figure}

The difference in the model's predictions to the true label for the range of epsilon values for each dataset is shown in the histogram plots, i.e. \cref{fig:fgsm_mnist_hist}, \cref{fig:fgsm_pathmnist_hist} and \cref{fig:fgsm_organmnist_hist}. 
For the MNIST dataset, see ~\cref{fig:fgsm_mnist_hist}, beyond values of $\epsilon=0.55$, class $8$ is predicted as the dominant class, which is also evident from ~\cref{fig:mnist_pymde} (increase in the number of \emph{blue points}). We also analyzed the relative change in volume of each cluster of the predicted class, see ~\cref{fig:rdr_mnist}. This indicated that, the volume of the cluster pertaining to class $8$ gradually increases with an increase in epsilon (the yellow color indicates large difference in values). This can be mapped to cluster $8$ attracting more and more malignant samples and eventually becoming the dominant predicted class label.
For the PathMNIST dataset, the class $2$ is the predominantly predicted class as seen from ~\cref{fig:fgsm_pathmnist_hist} for high values of $\epsilon$. As noticed from ~\cref{fig:rdr_pathmnist} the volume of each class cluster decreases and also remain nearly the same (the black regions for $\epsilon=0.75,1$) with an increase in epsilon as compared to class $2$. The ~\cref{fig:pathmnist_pymde} further indicates that for higher values of epsilon, the classes cluster into a confined space contributing to the increase of density well above $1$. On the other hand, the coverage decreases as the diversity in the generated AEs decreases. A similar trend to the MNIST dataset was also noticed in the case of the OrganMNIST dataset.

\begin{figure}
     \centering
     
        \subfigure[MNIST]{\label{fig:clust_mnist}{\includegraphics[width=\textwidth] {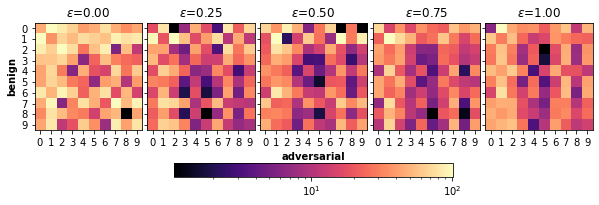}}}%
        \vfill
        \subfigure[PathMNIST]{\label{fig:clust_pathmnist}{\includegraphics[width=\textwidth] {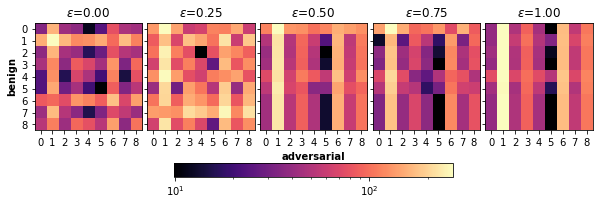}} }
        \vfill
        \subfigure[OrganMNIST]{\label{fig:clust_orgmnist}{\includegraphics[width=\textwidth]{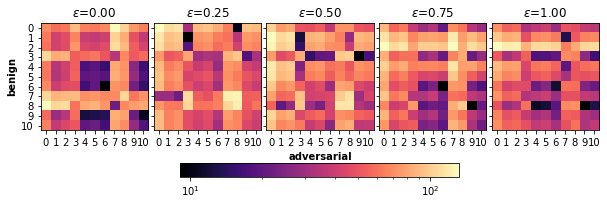}}}
    \caption{The difference in cluster centres between the classes for each epsilon. The y-axis indicates the class labels for benign samples and the y-axis for AEs. The variation of the distance matrices between epsilons indicate a change in cluster distance.}
   \label{fig:cluster_data}
\end{figure}

We also analyzed the change in distance between each cluster center between the benign samples and the generated AEs. The pair-wise distance between the benign and AEs were computed, which then served as input for Multi-dimensional scaling to obtain the reduced dimensional coordinates for the cluster centres. As seen from ~\cref{fig:clust_mnist} and~\cref{fig:clust_pathmnist}, the distance matrix varies for each epsilon value indicating an evident change in cluster centers for MNIST and PathmMNIST datasets. However, the ~\cref{fig:clust_orgmnist} shows that such onset in difference appears only after $\epsilon=0.5$ for the OrganMNIST dataset. These results also purport the variation of the metrics explained in ~\cref{fig:den_exp}.

\begin{figure}[ht]
    \centering
    \includegraphics[width=\textwidth]{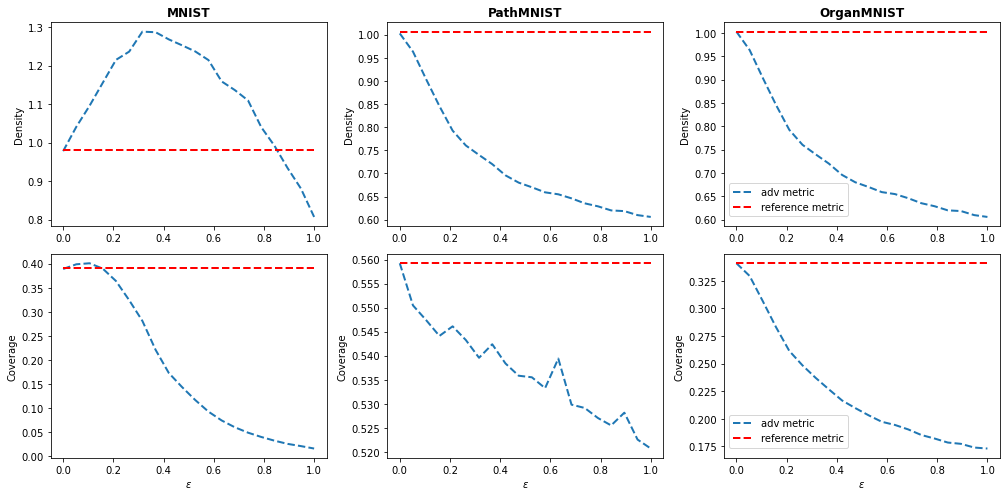}
    \caption{The variation of \dens{} and \covr{} for the reference validation dataset as well as for AEs using the embeddings. The red line corresponds to metrics obtained from the benign (reference) validation set. The blue line refers metrics obtained from malignant AEs derived from the same validation set. The green and gray bands indicate uncertainty intervals between $25\%-50\%$ and $50\%-75\%$ quantiles estimated from ensemble measurements respectively. Each column corresponds to the specified dataset as indicated. Rows report results for \dens{} or \covr{} respectively. \Dens{} and \covr{} from AEs are distinctly different from reference values.}
    \label{fig:fgsm_embed}
\end{figure}

The results of an analysis using the embeddings from the last layer of a classification model in contrast to using the input images are shown in ~\cref{fig:fgsm_embed}. There exists a perceptible difference between the reference and adversarial \dens{} and \covr{} for each of the datasets. We perceive this as a clear indication, that the morphology of the embedded space is severely different to that of the input space of raw images. However, we see that AEs can be differentiated from benign samples in this scenarios too albeit under different morphological constraints.

\clearpage

Similar to the FGSM attack, we performed the experiment of mixing AEs into the full batch of validation(benign) samples for the boundary attack. The variation in \dens{} and \covr{} values is shown in \cref{tab:bdy_mixture_den} and \cref{tab:bdy_mixture_cov}. These results re-iterate the characteristics of these metrics to distinguish batches of benign samples from adversarial samples.

\begin{table}[ht]
\centering
\setlength{\tabcolsep}{2pt}
\begin{tabular}{ccccc} 
    \toprule
    \textbf{Dataset} & \multicolumn{3}{c}{\textbf{Density}} \\ \midrule
        AE admixture & $0\%$ & $25\%$ & $50\%$ & $75\%$ \\ \midrule
        \textbf{MNIST} & $1.007$ & $0.80$ & $0.59$ & $0.39$ \\ \midrule
        \textbf{PathMNIST} & $0.83$ & $3.002$ & $4.92$ & $7.91$ \\ \midrule
        \textbf{OrganMNIST} & $1.01$ & $1.41$ & $1.86$ & $2.29$ \\
        \bottomrule
\end{tabular}
\caption{\Dens{} from batches of benign samples with proportionate admixtures of AEs for the boundary attack.}
\label{tab:bdy_mixture_den}
\end{table}

\begin{table}[ht]
\centering
\setlength{\tabcolsep}{2pt}
\begin{tabular}{ccccc} 
    \toprule
    \textbf{Dataset} & \multicolumn{3}{c}{\textbf{Coverage}}\\ \midrule
        AE admixture & $0\%$ & $25\%$ & $50\%$ & $75\%$\\ \midrule
        \textbf{MNIST} & $0.14$ & $0.11$ & $0.07$ & $0.04$ \\ \midrule
        \textbf{PathMNIST} & $0.49$ & $0.51$ & $0.56$ & $0.59$ \\ \midrule
        \textbf{OrganMNIST} & $0.34$ & $0.30$ & $0.27$ & $0.25$ \\
        \bottomrule
\end{tabular}
\caption{\Covr{} from batches of benign samples with proportionate admixtures of AEs for the boundary attack.}
\label{tab:bdy_mixture_cov}
\end{table}

\clearpage

\section{Runtime analysis} \label{runtime}
\begin{figure}[ht]
    \centering
    \includegraphics[width=\textwidth]{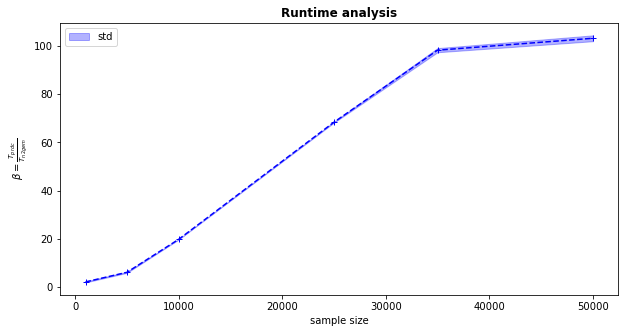}
    \caption{The speed up in runtime between \emph{prdc} and our implementation. Indicates the time for computing both \dens{} and \covr{} on a dataset with specified sample size and $512$ dimensions.
    }
    \label{fig:time}
\end{figure}

The analysis of runtime between \texttt{prdc} and our implementations is shown in ~\cref{fig:time}. We compute the speed up, $\beta=\frac{T_{prdc}}{T_{n2gem}}$, obtained in using our implementation in comparison to prdc. We have decreased the computation time by nearly $100x$ when compared to \texttt{prdc} implementation. 


 \clearpage
 
\section*{Review and Author response}
In this section, we provide our response to the questions and concerns brought forward by two reviewers during peer-review at the \href{https://advml-frontier.github.io/}{AdvML} workshop with \href{https://icml.cc/Conferences/2022/CallForPapers}{ICML 2022} where this preprint was submitted to. Let us first of all thank the reviewers for investing their time in looking through our article in detail.

\subsection*{Reviewer $1$}

\begin{enumerate}
    \item "It seems that the FGSM attack and MNIST dataset adopted in this paper could not reflect the effect of this
method, maybe presenting the performance on Cifar-10 under PGD attack will be more convincing." \\[12pt]
    \textbf{Our reply}:
    \begin{itemize}
        \item We have shown the results of the metrics \dens{} and \covr{} on the MNIST dataset. To improve the complexity of the attack and the dataset used, we have conducted experiments with $2$ additional and very different bio-medical datasets, OrganMNIST and PathMNIST. To complement, we also used a black-box attack and hoped that in light of the compute time expense that these two attacks provide enough spectrum on the kind of attack.
        \item The results indicated the capability of the metrics \dens{} and \covr{} to detect adversarial samples in all settings tested. The results in ~\cref{bb_atk} discuss these observations. 
        \item In particular, \cref{fig:fgsm_variation} addresses the concern by the reviewer. The text in ~\cref{bb_atk} discusses why we believe this to be evidence in favor of our method. 
    \end{itemize}
    
    \item "In 3.2, the Density metric is unbounded, and the Coverage is bounded between 0 and 1. why are these two
metrics different." \\[8pt] 
    \textbf{Our reply}:
    
    \begin{itemize}
        \item As cited in ~\cref{den}, the density metric measures the number of benign-sample manifolds containing an adversarial sample. Inherited in the formulation of density, the metric's value can surpass $1$, indicating that the benign-sample neighborhoods and the adversarial samples are very closely packed. Coverage, on the other hand, computes the fraction of benign-sample hyperspheres containing at least one adversarial sample. Both metrics have very different meanings, and in generative modeling, density demonstrates the quality and coverage of the variability of the image for a given dataset. 
    \end{itemize}

    \item "Through the experiments, under FGSM, in Figure 3 (a) and Figure4, why does PathMNIST express different
trends following the changing of epsilon?" \\[12pt]
    \textbf{Our reply}:
    \begin{itemize}
        \item The causes for the variation of \dens{} and \covr{} have been discussed in ~\cref{fgsm}. Furthermore, dimensionality reduction techniques also describe the reason for such particular trends. The related analysis and results are shown in the ~\cref{Met_analysis}.
    \end{itemize}

    \item "Figure 3 should be modified to more clear, the legend, x/y-axis, and the line in the figure are not clear." \\[12pt]
    \textbf{Our reply}:
    \begin{itemize}
        \item This is a valid statement. We could include an explanation of $\epsilon$ which is given in the main text, but not in the caption. 
        \item We could add in the plot legends the word "quantiles" (e.g. $25\%-50\%\,\text{quantiles}$) to repeat the caption in the plot. We agree that this may lead to a more quick visual digestion.
    \end{itemize}
    
\end{enumerate}

\subsection*{Reviewer $2$}
\begin{enumerate}
    
    \item "The chosen dataset is too simple and should be tested on at least the CIFAR-10 dataset" \\ [12pt]
    \textbf{Our reply}:
    \begin{itemize}
        \item We focussed our introductory study on the MNIST dataset, which we acknowledge as a valuable benchmark dataset to validate a few claims. However, we have experimented with and validated our proposed metrics on $2$ very different bio-medical datasets, which we believe are on par with the CIFAR-10 dataset.
        \item In addition, we had hoped (but not documented in the text due to space constraints) to explicitly not include too many established benchmark datasets, as various authors \citet{carter2021overinterpretation, donewithimagenet} have shown problems of overfitting on these in publications.
        \item Last, the workshop website did not mention that using CIFAR-10 as a training dataset is a requirement for a submission.
    \end{itemize}
    
    \item "Why are medical datasets selected? In other words, I would like to know in what scenarios medical images are
attacked and need to be identified as adversarial examples" \\ [12pt]
    \textbf{Our reply}:
    \begin{itemize}
        \item DNNs have been increasingly used in medical diagnostics and as a tool for physicians in intricate fields of medicine such as radiology and ophthalmology. Trillions of dollars worth of insurance claims are processed by companies leveraging such automated systems. Not only focussing on financial incentives but the vulnerability of algorithms should also be considered in such a high stake field~\citep{finlayson2019adversarial}. We considered this situation a ground model, concentrated on medical images, and diversified our validation approach. 
    \end{itemize}
    \item "Limited novelty. For the problem this paper is trying to solve, the most critical is metric, which has been proposed
by Naeem et al." \\ [12pt]
    \textbf{Our reply}:
    \begin{itemize}
        \item The \href{https://advml-frontier.github.io/}{AdvML2022 website} clearly invites articles on "Adversarial ML metrics and their interconnections"
        \item To our knowledge, \dens{} and \covr{} are metrics conceived for characterizing the performance of GANs~\citep{naeem2020reliable}. 
        \item As stated in ~\cref{sec:intro}, the novelty of our approach lies in the mapping of metrics from the field of generative modeling to adversarial attacks. 
        \item Moreover, we have also worked to improve the runtime of the metrics in comparison to the previous implementation, see \cref{runtime}.
    \end{itemize}
    
    \item "The validity of the proposed method needs to be verified with the latest proposed attack" \\ [12pt]
    \textbf{Our reply}:
    \begin{itemize}
        \item In our study, we have experimented with both the extremes of adversarial attacks, i.e., white- and black-box attacks.  We believe that the proposed metrics would scale well in detecting adversarial samples as our approach remains model agnostic and only relies on the datasets in hand and the generated adversarial samples.
        \item Similar to our reply above, this appears a requirement to us which is not stated on the \href{https://advml-frontier.github.io/}{AdvML2022 website}.
    \end{itemize}
\end{enumerate}
\end{document}